# Automated HIV Screening on Dutch Electronic Health Records with Large Language Models

Lang Zhou[1,2], Amrish Jhingoer[1,2], Yinghao Luo[1,2], Klaske Vliegenthart–Jongbloed[3,4], Carlijn Jordans[4], Ben Werkhoven[5], Tom Seinen[6], Erik van Mulligen[6], Casper Rokx[3,4], and Yunlei Li[1]⋆

[1]Department of Pathology & Clinical Bioinformatics, Erasmus University Medical Center Rotterdam
[2]Department of Computer Science, Vrije Universiteit Amsterdam
[3]Department of Internal Medicine, Erasmus University Medical Center Rotterdam
[4]Department of Medical Microbiology & Infectious Diseases, Erasmus University Medical Center Rotterdam
[5]Department of Data & Analytics, Erasmus University Medical Center Rotterdam
[6]Department of Medical Informatics, Erasmus University Medical Center Rotterdam

**Abstract.** Efficient screening and early diagnosis of HIV are critical for reducing onward transmission. Although large-scale laboratory testing is not feasible, the widespread adoption of Electronic Health Records (EHRs) offers new opportunities to address this challenge. Existing research primarily focuses on applying machine learning methods to structured data, such as patient demographics, for improving HIV diagnosis. However, these approaches often overlook unstructured text data such as clinical notes, which potentially contain valuable information relevant to HIV risk. In this study, we propose a novel pipeline that leverages a Large Language Model (LLM) to analyze unstructured EHR text and determine a patient's eligibility for further HIV testing. Experimental results on clinical data from Erasmus University Medical Center Rotterdam demonstrate that our pipeline achieved high accuracy while maintaining a low false negative rate.

## 1 Introduction

### 1.1 Background

Human Immunodeficiency Virus (HIV) attacks the body's immune system and is the direct cause of Acquired Immunodeficiency Syndrome (AIDS). The United Nations estimates that 1 out of 200 people in the world are infected with HIV in 2023 [1]. Efficient screening of HIV is a crucial topic in the public health sector, essential for early intervention and reducing transmission rates.

---

⋆ y.li.1@erasmusmc.nl

In clinical practice, HIV screening primarily occurs through voluntary testing initiated by patients or recommendations from healthcare providers when specific symptoms or risk factors are identified. Despite substantial efforts, many HIV-positive individuals remain undiagnosed due to factors such as stigma, lack of awareness, or perceived low-risk status, delaying treatment initiation and increasing the risk of transmission. Consequently, relying solely on patient-initiated testing or symptom-based screening is insufficient for comprehensive detection and containment of HIV.

To address the limitations of current screening practices, efforts have been made to develop universal guidelines for HIV testing recommendations. The EuroTEST guideline [2] offers a standardized framework for determining HIV testing eligibility, based on a predefined set of indicator conditions, clinical presentations, and exclusion criteria. By providing clear rules, the guideline enables consistent and effective decision-making across healthcare settings. However, thoroughly reviewing patients' health records based on such guidelines requires significant manual effort from healthcare providers, leading to increased workloads and potential human errors, thus creating a demand for more efficient and automated approaches.

## 1.2 Our Contribution

In this research, we propose an automated pipeline which analyzes Dutch EHR data and makes decisions regarding the necessity of further HIV testing for patients. Our approach utilizes MedGemma [3][4], a large language model (LLM) developed by Google and pretrained extensively on medical texts. To ensure alignment with established clinical guidelines, we incorporated the EuroTEST guideline directly into the prompt, enabling the system to make decisions consistent with current best practices in HIV screening.

This paper also aims to address the following research questions, the answers to which may provide valuable insights for the practical deployment of the proposed pipeline:(1) Does a complex prompt with more detailed guidelines lead to improved pipeline performance compared to a simple prompt with general guidelines? (2) What is the most effective method for obtaining a reliable prediction when multiple runs of the model are performed? (3) Are there observable correlations between the lengths of the EHR input, model output and prediction correctness?

## 2 Relevant Studies

### 2.1 Automated HIV Screening

Automated screening of HIV based on patients' biomedical data has been widely researched. Previous research focuses mainly on the use of structured data. Ahlström et al. [5] performed logistic regression on demography and hospital diagnosis data from Danish civil registry databases, achieving an AUC-ROC score of 0.884 on the validation data. Saha et al. [6] proposed an optimized ensemble learning model for predicting HIV based on patient background and behavioral data, yielding significantly better accuracy compared to that of a single model.

With the advance in NLP techniques, various researches also investigate using unstructured text information, especially EHR data. Feller et al. [7] utilized Latent Dirichlet Allocation (LDA) [8] for automated keyword identification and topic modeling, integrating text-based features with structured data. This approach yielded an F1 score of 73.3%, outperforming the baseline model, which relied solely on structured information and achieved an F1 score of 59.2%. In recent years, language models based on the Transformer architecture [9] have become increasingly popular in the NLP field due to their great linguistic understanding capability and high computing efficiency. Sánchez et al. [10] employed RoBERTa$_{Bio}$ [11], a Transformer-based Pretrained Language Model (PLM) trained on Spanish biomedical texts, to perform binary classification on suspicious versus non-suspicious HIV cases. Compared to a traditional NLP method which used TF-IDF vectorization and a max-entropy classifier, the PLM method achieved a superior F1 score as well as a lower false negative rate.

## 2.2 Dutch Language Models for Biomedical Tasks

This research project focuses on analyzing EHR in Dutch, highlighting the need for a language model with Dutch language understanding capability as well as specialty in the medical domain. Verkijk et al. [12] proposed MedRoBERTa.nl, a Dutch PLM based on the RoBERTa architecture [13], pretrained on 13 GB of Dutch EHR data from Medical Centre of the Vrije Universiteit (VuMC) and the Amsterdam Medical Centre (AMC). Evaluation of biomedical text embedding accuracy demonstrated that MedRoBERTa.nl consistently outperformed general-purpose Dutch PLMs such as BERTje [14], which was trained on Wikipedia and news data.

## 2.3 Gaps in Current Research

Despite progress in automated HIV screening, important gaps remain. Most approaches still rely on structured data and overlook the rich information in unstructured clinical notes [5][6][7]. Several studies [10][12] have applied transformer-based models for text understanding, but these have not extended to the use of more recent generative large language models.

To date, no published work has investigated generative LLMs for automated HIV screening. Generative LLMs offer several advantages: they can reason over long, complex clinical narratives; provide human-readable reasoning output; and adapt more flexibly to the heterogeneous documentation styles commonly found in real-world EHRs. Addressing this gap could capture clinical context overlooked by traditional approaches and significantly improve early detection performance.

## 3  Methodology

### 3.1  Overview

Figure 1 gives an overview of our proposed pipeline. The Erasmus HIV dataset is first divided into a training set (90%) and a testing set (10%). The testing set is used solely for evaluation purposes during this research.

For the LLM, an evaluation pipeline is constructed to assess its classification capability. For each instance in the testing set, a prompt is generated, instructing the MedGemma model to make a clinical decision regarding the necessity of HIV testing. These prompts simulate real-world clinical decision support queries. The generated textual responses are then parsed to extract binary decisions, which are compared against ground truth labels.

In parallel, the training set undergoes a downsampling procedure to address class imbalance, resulting in a balanced subset. This subset is used to fine-tune a multilingual BERT model (mBERT) [15], producing a task-specific classifier. The classifier is then evaluated on the reserved testing set, and its performance serves as the baseline for comparison.

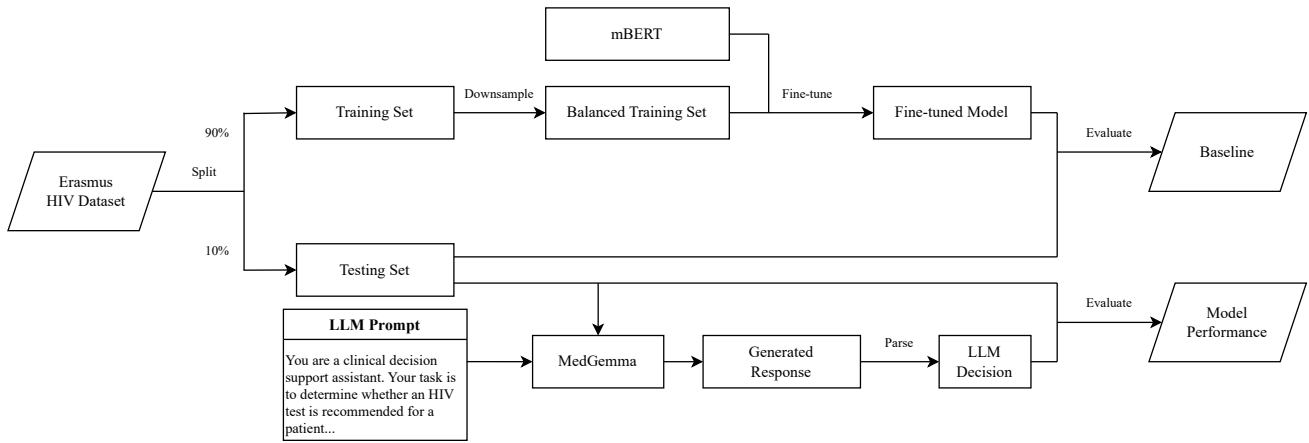

Fig. 1: Overview of the pipeline

### 3.2  Erasmus HIV Dataset

The Erasmus HIV Dataset [16] is a binary text classification dataset created from internal EHR data of Erasmus University Medical Center Rotterdam (Erasmus MC). The cohort is part of the project #aware.hiv and was approved under MEC-2020-0140. Patients aged 18 and older, without a known HIV diagnosis but with a registered HIV indicator condition between 1-1-2020 and 31-7-2023 are included. The classification targets are `Exclusion` (patient should be excluded from further HIV testing, represented by the binary label 0) and `Inclusion` (patient should be included for further HIV testing, represented by the binary label 1). The manually curated class labels were created by an HIV expert according to EuroTEST. Patient IDs were pseudonymized before data extraction.

**Data Generation & Preprocessing** Since EHR data and clinical decision labels are stored in different systems, we need to merge the obtained two raw data files first. Table 1 shows a description of the raw data fields.

Table 1: Description of raw data fields.

| Type | Field | Description |
|---|---|---|
| Free Text | authored | Generation date of the corresponding EHR entry |
| | section_text | EHR text |
| ——— | Pseudoniem | Unique numerical ID for each patient, anonymized |
| Metadata | start_date | Generation date of the diagnosis |
| | icd10_code | Automatically flagged ICD-10 codes by Erasmus MC's system |
| | specialism | Source department of the diagnosis |
| | HIV_indicator_HIVteam | Decision flag; 0 for exclusion, 1 for inclusion, 2 for not selected for research |

To merge the free-text and metadata, we first aggregated the EHR free-text entries by grouping them based on `Pseudoniem` in descending order of time, resulting in a concatenated EHR text for each patient. Following aggregation, the free-text data were merged with the metadata by matching entries on the `Pseudoniem` field. The final merged dataset includes each patient's `Pseudoniem`, the aggregated EHR free text, and the corresponding decision label. Entries with a decision label of 2 were excluded, as they were not selected by the clinical team for this study.

The text data were pre-anonymized by Erasmus MC's system. To further ensure data quality, we applied additional preprocessing by removing escape characters (e.g., `\n`) and eliminating any instances of mojibake.

After generating the full dataset, we randomly sampled 10% of the full data to create the test set. Stratified sampling was applied to ensure that the class distribution in the test set reflects that of the full dataset, thereby maintaining a representative balance between inclusion and exclusion cases.

**Incorporation of Structured Information** In the later stage of this study, we also gained access to patients' prescribed medication and virology test data, providing the model with a more comprehensive view of each patient's clinical profile. The relevant fields of the structured data are described in Table 2.

For each patient in the Erasmus HIV dataset, we first retrieved the associated structured data. These data were then transformed into a textual format using the scheme illustrated in Figure 2. The resulting structured text was appended to the corresponding free-text EHR entry to form a unified input for the model. In cases where no structured information was available, placeholder values (e.g., None) were inserted in place of the structured sections to maintain format consistency.

Table 2: Description of structured data fields

| Type | Field | Description |
|------|-------|-------------|
| Medication | code5_ATC_code | ATC code for the medication |
| | code_text | Medication name |
| | Pseudoniem | Unique numerical ID for each patient, anonymized |
| Virology Test | hix_code | Name of virology test |
| | valueString | Test result string |

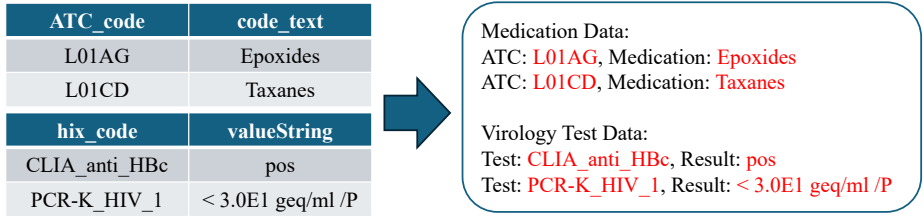

Fig. 2: Illustration of structural to textual transformation

**Statistics** The full Erasmus HIV Dataset contains 10626 entries. After splitting, the train set has 9563 entries with 8614 excluded and 949 included cases, the test set has 1063 entries with 957 excluded and 106 included cases.

### 3.3 Model

**MedGemma** MedGemma is an LLM developed by Google and released in May 2025. Built on the Gemma 3 architecture, MedGemma is further pretrained on a large-scale corpus of medical texts. It is available in three variants: a 4B or 27B multi-modal version capable of processing both text and images, and a 27B text-only version. As this study focuses exclusively on textual data, we use the 27B text-only variant.

We selected MedGemma over other models due to its combination of strong Dutch language understanding capability and extensive pretraining on medical content. Its base model, Gemma 3, is pretrained on a multilingual corpus of 14 trillion tokens [17] and demonstrates competitive performance on the EuroEval Dutch benchmark [18][19]. Furthermore, performance evaluations conducted by Google [4] indicate that MedGemma substantially outperforms the base model on a range of medical benchmarks following its domain-specific pretraining on medical data.

**mBERT** First introduced in 2018 [15], mBERT is a PLM trained on Wikipedia data from 104 languages, including Dutch. It is selected as the baseline in this study due to its widespread use in similar medical NLP research [20][21][10].

We used the `bert-base-multilingual-cased` variant of mBERT, which consists of 179 million parameters and is case-sensitive. To adapt the model to the Dutch medical domain, we further pretrained it on the free-text data from the Erasmus HIV dataset.

### 3.4 Experiment Setup

**Prompt** We constructed the system prompt based on the EuroTEST guideline. The prompt defines the model's role as a clinical decision support assistant, tasked with assessing whether an HIV test is recommended for a given patient based on a Dutch-language medical note. The core criterion is that an HIV test is warranted if at least one of the 36 recognized HIV indicator conditions (ICs) is present, and no valid exclusion criteria apply. Chain-of-Thought (CoT) reasoning structure is integrated to enhance reasoning performance and decision transparency.

Two prompt variants were used in the experiment: Simple Prompt (SP) and Complex Prompt (CP). The only difference between the two prompts is that SP uses a 3-step CoT while CP uses a 9-step one. The CoT of CP is an extension based on that of SP, with more guidelines on the interpretation of the structural information, exemption criteria, and a more sophisticated decision rule. To ensure consistency, the rest of the prompts remain identical.

Both prompts were developed in close collaboration with the clinical team to ensure their consistency with the guideline and clinical practice. The prompts are available in the Appendices A.1 and A.2.

**LLM Inference** To strike a balance between model performance and efficiency, we used the `medgemma-27b-text-it-UD-Q8_K_XL` variant, an 8-bit dynamically quantized version of the MedGemma model provided by Unsloth [22]. The model was hosted using the `llama.cpp` library on a server equipped with a single NVIDIA H100 Tensor Core GPU.

Table 3 shows the inference configuration we used. The choice of these parameters was based on official recommendations from the Gemma team [23]. To control the creativity and variability of the generation process, the temperature parameter was reduced from the default value of 1.0 to 0.8.

For each record in the test set of the Erasmus HIV dataset, we first constructed the model input using the predefined prompt template. The input was then processed by the model in three independent runs. For each run, we recorded the generated output and the predicted decision label. Since we specifically asked the model to conclude with a YES or NO token at the end of the output, the decision label was directly obtained from a keyword search within the last 10 generated tokens.

Table 3: Inference configuration for MedGemma

| Parameter | Value |
|---|---|
| temperature | 0.8 |
| top_k | 64 |
| top_p | 0.95 |
| min_p | 0.0 |

**Output Aggregation Strategies** To enhance decision stability and the performance of the LLM, we explored two output aggregation strategies: Best-of-N sampling [24] and self-consistency [25]. These methods aim to reduce variability and improve alignment with clinical guidelines by aggregating multiple outputs.

Best-of-N sampling refers to generating multiple outputs for a given input, and selecting the best output based on a confidence metric or reward function. In this study, we use average log-probability (AvgLogP):

$$\text{AvgLogP} = \frac{1}{n} \sum_{i=1}^{n} \log p(y_i \mid x, y_{<i}) \qquad (1)$$

where $n$ is the number of generated tokens, $p(y_i \mid x, y_{<i})$ is the model's predicted probability for the generated token $y_i$ given the input $x$. A higher AvgLogP reflects that the model consistently assigns higher probability scores to its generated tokens, indicating greater confidence.

As a simpler alternative, self-consistency decoding uses a majority voting mechanism to find the most reliable output, eliminating the need for explicit scoring.

Additionally, we included four more output selection methods:

1. **First Prediction.** Select the first prediction among three runs, which simulates the situation where no output aggregation is performed.
2. **Shortest Output.** Select the prediction with least output tokens.
3. **Longest Output.** Select the prediction with most output tokens.
4. **No Inconsistent.** Discard the entry when three runs do not give the exact same decision, leaving the case for manual review.

**Fine-tuning of mBERT** To fine-tune the mBERT baseline, we first balanced the training set to avoid overfitting and training instability. The majority class (excluded cases) was under-sampled to maintain a $1 : 1$ ratio with the included class, yielding a training set with 1898 records.

The implementation of fine-tuning was through the `SimpleTransformers` library [26]. Table 4 shows the hyper-parameters.

Table 4: Hyper-parameter for fine-tuning of mBERT

| Parameter | Value |
|---|---|
| learning rate | 2e-5 |
| epoch | 3 |
| optimizer | AdamW |
| stride | 0.8 |

The maximum context window of mBERT is 512 tokens. To handle records longer than this value, a sliding window approach was applied. Each record was divided into sub-sequences no longer than 512 tokens, with each sub-sequence inheriting the original record's label. The model was then trained on this set of sub-sequences. To avoid information loss between sub-sequences, the stride parameter was set to 0.8 so that the window would be slid for 80% of the maximum context window before obtaining the next sub-sequence, maintaining a 20% overlap between adjacent sub-sequences. During inference, the model would predict on each sub-sequence, and the prediction for the full record would be the mode of predictions from all the sub-sequences.

**Evaluation Metrics** In this research, we evaluate model performance using three primary metrics: **Accuracy**, **Macro-F1**, **Sensitivity** and **Specificity**.

**Accuracy** measures the overall correctness of the model by calculating the ratio of correctly predicted instances to the total number of instances. Formally, accuracy is defined as:

$$\text{Accuracy} = \frac{\text{Number of correct predictions}}{\text{Total number of predictions}}$$

**Macro-F1** evaluates the model's performance across different classes independently. It calculates the F1 score for each class separately, then computes their unweighted average. The formula for Macro-F1 is:

$$\text{Macro-F1} = \frac{1}{C} \sum_{c=1}^{C} \frac{2 \cdot \text{Precision}_c \cdot \text{Recall}_c}{\text{Precision}_c + \text{Recall}_c}$$

where $C$ denotes the total number of classes, with precision and recall defined for each class $c$ as:

$$\text{Precision}_c = \frac{TP_c}{TP_c + FP_c}, \quad \text{Recall}_c = \frac{TP_c}{TP_c + FN_c}$$

Here, $TP_c$, $FP_c$, and $FN_c$ represent true positives, false positives, and false negatives, respectively, for class $c$.

**Sensitivity**, or Recall for the `Inclusion` (positive) class, specifically evaluates how effectively the model identifies Inclusion cases. It measures the proportion of correctly identified positive cases among all positive cases. Formally, sensitivity is defined as:

$$\text{Sensitivity} = \frac{TP}{TP + FN}$$

where $TP$ is the number of true positives (correctly predicted Inclusion cases), and $FN$ is the number of false negatives (Inclusion cases incorrectly predicted as Exclusion).

**Specificity**, also known as the *True Negative Rate*, evaluates how effectively the model identifies Exclusion (negative) cases. It measures the proportion of correctly identified negative cases among all actual negative cases. Formally, specificity is defined as:

$$\text{Specificity} = \frac{TN}{TN + FP}$$

where $TN$ is the number of true negatives (correctly predicted Exclusion cases), and $FP$ is the number of false positives (Exclusion cases incorrectly predicted as Inclusion). High specificity indicates that the model effectively avoids falsely classifying Exclusion cases as Inclusion.

In this research, we would focus on using sensitivity as the primary evaluation metric. For the HIV screening task, we aim to miss as few cases as possible. While a higher sensitivity may lead to lower precision or accuracy, we believe that the cost of a false positive (unnecessary test) is always less severe than that of a false negative (missed diagnosis).

## 4  Results

### 4.1  Results Overview

Table 5: Performance of MedGemma using different prompts and output aggregation methods, compared with mBERT baseline.

| Model | Method | Accuracy | Macro-F1 | Sensitivity | Specificity |
|-------|--------|----------|----------|-------------|-------------|
| MedGemma-SP | First Prediction | 67.26 | 56.27 | 85.85 | 79.10 |
| | Self-consistency | **69.33** | **58.18** | **88.68** | **81.71** |
| | Max-probability | 68.39 | 57.12 | 85.85 | 79.31 |
| | Shortest Output | 68.11 | 56.79 | 84.91 | 80.88 |
| | Longest Output | 67.26 | 56.27 | 85.85 | 78.16 |
| | No-inconsist | 74.93 | 65.77 | 90.00 | 87.65 |
| MedGemma-CP | First Prediction | 78.08 | 62.59 | **68.87** | 65.20 |
| | Self-consistency | **80.06** | **63.75** | 65.09 | **67.19** |
| | Max-probability | 77.80 | 61.55 | 64.15 | 66.46 |
| | Shortest Output | 79.30 | 63.05 | 65.09 | 66.25 |
| | Longest Output | 76.86 | 60.91 | 65.09 | 65.20 |
| | No-inconsist | 85.82 | 70.76 | 69.51 | 72.70 |
| mBERT | - | 81.65 | 67.03 | 75.47 | 82.34 |

The detailed results are presented at Table 5. Among the standard approaches, MedGemma-SP combined with the self-consistency method performed the best, achieving the highest sensitivity (88.68%) along with improvements in accuracy and macro-F1 compared to other output selection strategies applied to MedGemma-SP.

MedGemma-CP demonstrates improvements in accuracy and macro-F1 relative to MedGemma-SP, indicating enhanced overall decision quality. However, its sensitivity is consistently lower across all selection methods, with the highest value (68.87%) observed under the first-prediction strategy.

The baseline mBERT outperforms MedGemma-CP across all metrics, yet falls short of MedGemma-SP in terms of sensitivity, suggesting its limited ability to identify inclusion cases.

Lastly, the No-inconsist method yields the highest sensitivity for both prompts (90.00% for MedGemma-SP, 69.51% for MedGemma-CP), alongside top performance in other metrics. However, as it filters out inconsistent responses and thus removes certain data points, its results should be interpreted separately.

The complete confusion matrices are available in Appendix A.3 for detailed inspection.

### 4.2 Input, Output & Correctness Correlation

In this section, we present the results addressing Research Question 3. To ensure the validity of the analysis, we apply the $3\sigma$ outlier-removal rule, removing all token length[1] values that fall outside the range $[\mu - 3\sigma, \mu + 3\sigma]$.

Since all data groups fail the Shapiro-Wilk normality test [27] ($p < 0.05$), we would use non-parametric statistical tests to assess significance throughout this section.

Figure 3 visualizes[2] the correlation between input token lengths and average output token lengths. For MedGemma-SP, we obtain a Spearman rank correlation coefficient [28] of 0.613 ($p = 1.69 \times 10^{-110}$). For MedGemma-CP, we obtain a coefficient of 0.399 ($p = 8.33 \times 10^{-42}$).

Figure 4 visualizes the distribution of input token lengths grouped by prediction correctness. The Mann-Whitney U test [29] is performed to check for the difference between input token lengths for correct and incorrect predictions. For MedGemma-SP, using a two-sided alternative hypothesis, we obtain $p = 0.174$, indicating no significant difference in input lengths between the two groups. In contrast, for MedGemma-CP, with a one-sided alternative hypothesis testing whether correct predictions have longer input lengths, the test yields $p = 2.96 \times 10^{-3}$, suggesting a statistically significant difference.

Figure 5 visualizes the distribution of average output token lengths grouped by prediction correctness. Similarly, we use the Mann-Whitney U test to check for the difference between average output token lengths for correct and incorrect predictions. For MedGemma-SP with a one-sided alternative hypothesis testing whether correct predictions have shorter average output lengths, we obtain $p = 9.30 \times 10^{-3}$, indicating statistical significance. For MedGemma-CP using a two-sided alternative hypothesis, we obtain $p = 0.928$, suggesting no significant differences in output lengths.

---

[1] Token length is calculated through averaging the token lengths of outputs from three runs.

[2] Visualization is presented in log-scale for readability. Statistical test is conducted on the raw data.

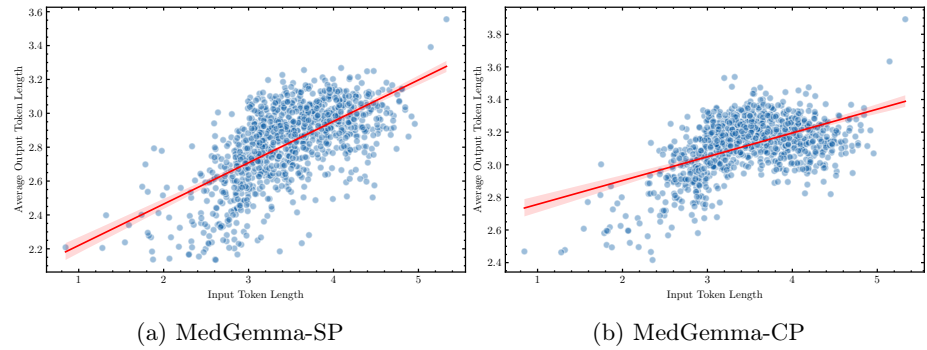

(a) MedGemma-SP                    (b) MedGemma-CP

Fig. 3: Relationship between input and output token lengths

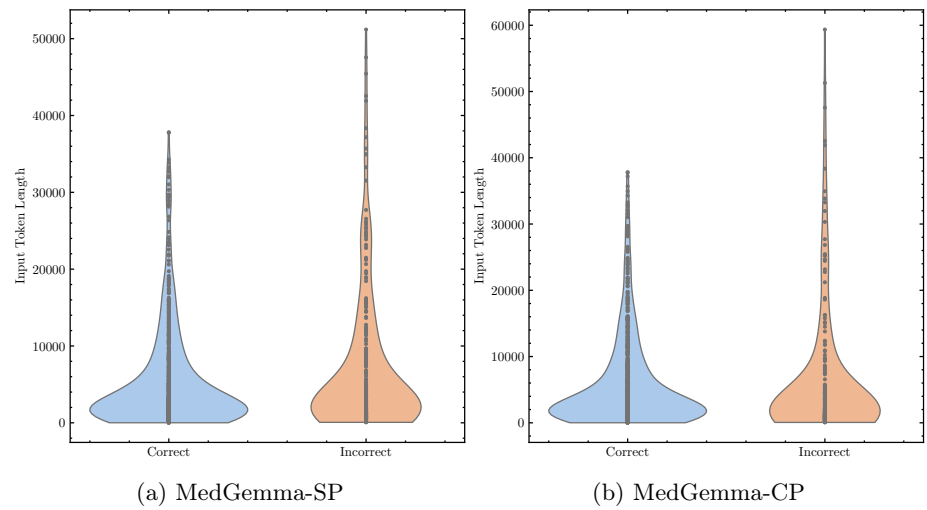

(a) MedGemma-SP                    (b) MedGemma-CP

Fig. 4: Distribution of input lengths grouped by correctness

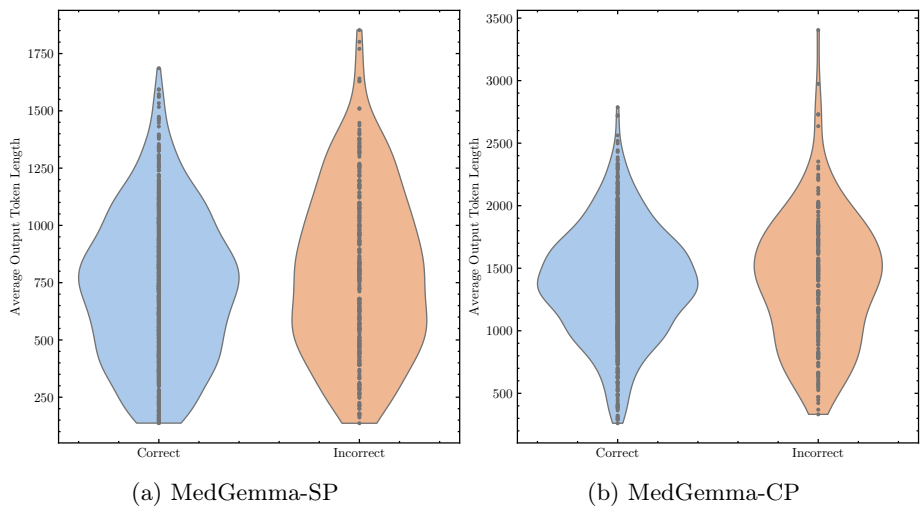

(a) MedGemma-SP          (b) MedGemma-CP

Fig. 5: Distribution of input lengths grouped by correctness

## 5  Discussions

### 5.1  Simple vs. Complex Prompt

Contrary to intuitive belief and our original expectations, MedGemma-CP with a more sophisticated prompt actually underperforms relative to MedGemma-SP in terms of sensitivity. While overall metrics such as accuracy and macro-F1 show improvement, the substantial drop in sensitivity indicates a higher risk of failing to identify undiagnosed cases. Given the clinical context of this study, where minimizing false negatives is critical, this reduction in sensitivity is a significant drawback and makes the complex prompt unsuitable for real-world deployment.

The unsatisfying result of CP also suggests that a long 9-step CoT might not be the optimal choice for the HIV screening application. Although it is commonly assumed that longer CoTs would typically enhance model performance [30], this assumption has been increasingly questioned in recent literature. [31] provides a theoretical framework demonstrating that an optimal CoT length exists for each task-model pairing. Their experimental results indicate that this optimal length tends to decrease as model size increases. For example, in experiments with the Qwen2.5 series, the optimal CoT length was found to be 8 for the 7B model and 5 for the 32B model. Based on this trend, the 9-step CoT used with the 27B MedGemma model in our study may exceed the optimal reasoning length, potentially leading to degraded performance.

To better understand the potential reasons behind longer CoT's poor performance, we look back into the model outputs. The average output token length for MedGemma-SP is 751, while for MedGemma-CP it is 1403. Dividing these totals by the respective number of CoT steps, we obtain a rough estimate[3] of 250 average reasoning tokens per step for SP and only 155 for CP. This shows that while CP uses a CoT with three times the length of SP, the reasoning length for each step actually decreases by approximately 40%, suggesting shallower reasoning at each CoT step, which could negatively impact the model's decision quality.

Based on the discussion above, we conclude that using a complex prompt with more detailed guidelines does not necessarily lead to improved pipeline performance compared to a simple prompt with general guidelines. However, the complex prompt still has its own advantages in terms of overall decision accuracy and enhanced clinical interpretability. Future work may benefit from refining the current prompt versions to build an optimal prompt that strikes a better balance between reasoning depth and output quality.

### 5.2 Prediction Robustness

It can be observed from Table 5 that among the standard output selection methods, the Self-consistency method consistently yields the best overall decision quality for both prompts. The No-inconsist method, which filters out all entries with inconsistent predictions across runs, achieves the best performance across all metrics, at the cost of reducing the evaluation set by approximately 35% for SP and 24% for CP.

Additionally, we notice that the Max-probability method, which takes the prediction with the highest average log-probability, produces suboptimal results. This suggests that a high log-probability score, often interpreted as a measure of model confidence, does not necessarily correspond to prediction reliability in this context.

We conclude that both the Self-consistency and the No-inconsist methods demonstrate promising performance for the HIV screening task. The choice between them should be guided by the intended deployment scenario of the pipeline. If the goal is to develop a fully automated flagging system with minimal human intervention, the Self-consistency method is more appropriate, as it achieves high decision quality without discarding any input entries. On the other hand, if the system is designed as a human-AI hybrid, where clinician experts can review cases with inconsistent predictions, the No-inconsist method becomes more suitable due to its superior sensitivity. Furthermore, the excluded entries in the No-inconsist method can serve as valuable case studies for refining the prompt and improving pipeline robustness in future iterations.

---

[3] This estimation does not account for variation in the complexity of each CoT step or tokens unrelated to the reasoning process.

## 5.3 Correlations Between Lengths and Correctness

Regarding input lengths, our findings indicate that longer EHR inputs positively correlate with prediction correctness for the MedGemma-CP model. This suggests that MedGemma-CP benefits significantly from more context-rich information, which aligns with expectations since the complex prompt contains more detailed instructions on handling input information. Conversely, this correlation was absent for the MedGemma-SP model, implying that the simpler prompt does not leverage additional input complexity as effectively. A possible explanation is that MedGemma-SP's simpler structure does not fully exploit the rich details provided by lengthier inputs, thus leading to a less significant impact on prediction accuracy.

Analyzing model output lengths yields contrasting results. For MedGemma-SP, correct predictions significantly correlate with shorter outputs, suggesting output conciseness as an indicator of high-quality decision-making. This may imply that when MedGemma-SP makes correct predictions, it efficiently identifies critical decision factors without redundant elaboration. Oppositely, when the output from MedGemma-SP is lengthy, it could possibly be an indicator of hallucination or over-explaining due to uncertainty, rather than being a sign of quality reasoning. However, no significant relationship between output length and correctness was observed for MedGemma-CP. It can be explained that since the complex prompt has a longer CoT, the model would consistently produce thorough and detailed outputs regardless of correctness, weakening the discriminatory power of output length as a correctness indicator.

Overall, our results indicate that the correlation between prediction correctness and the lengths of the EHR input and model output is prompt-dependent. Longer input texts tend to improve prediction correctness when using a complex prompt, while shorter output lengths are associated with more accurate predictions under a simpler prompt setting. These findings highlight the potential utility of input and output lengths as indirect indicators of model reliability. Future work could explore practical applications of such correlations. For example, incorporating length-based features into decision confidence estimation, or developing a hybrid decision system in which entries with abnormal lengths would be flagged for manual review. Such strategies may enhance the robustness and safety of real-world deployment.

## 5.4 Limitations

This study has several limitations, mainly due to constraints in time and available resources.

Firstly, our analysis did not examine the model's reasoning outputs in detail to identify specific steps or details that the model overlooked, leading to incorrect predictions. Insights from such analysis could be used to refine the prompt iteratively.

Secondly, the EHR data used in our experiments were exclusively sourced from Erasmus MC. This single-source dataset could introduce biases specific to the institution's writing styles or clinical documentation conventions. Expanding the dataset to include EHRs from multiple institutions would strengthen the generalizability and robustness of our findings.

Lastly, due to strict offline inference constraints, we were unable to compare MedGemma's performance with larger state-of-the-art reasoning models available exclusively via commercial APIs, such as OpenAI's o3 [32], which currently leads several medical NLP benchmarks [3]. We also did not evaluate open-source alternatives such as DeepSeek-R1 [33] due to limited computing resources. Such comparative analyses would provide valuable insights into the relative strengths and weaknesses of MedGemma in clinical decision-support tasks.

### 5.5 Future Work

Based upon the findings and limitations identified in this study, we suggest the following directions for future work.

**Curated Dataset** Manual validation by clinicians should be conducted on the Erasmus HIV dataset to improve the quality of labels, thus enhancing evaluation reliability. Preferably, adding manual annotations of CoT would make model fine-tuning or few-shot prompting possible.

**Model Ensembling** Currently, our solution for integrating structural information involves transforming structured data into text and appending it to the input. However, structural information often follows simple, rule-based logic, which could be effectively handled by traditional machine learning models. Investigating an ensemble approach, which combines LLM-based models for free-text analysis and rule-based models for structured data, might enhance the robustness and explainability of the pipeline.

**Pipeline Deployment** This study did not extensively address the practical deployment of the proposed pipeline. Future work should explore critical deployment considerations, including integration into clinical workflows, compliance with clinical standards, and adherence to ethical and regulatory requirements.

## 6 Conclusions

In this research, we propose a pipeline that utilizes the MedGemma-27B model for automated HIV screening on Dutch EHR data. Evaluation on internal clinical data from Erasmus MC shows that the pipeline achieves a low false negative rate while maintaining strong overall decision accuracy. We believe that the findings from this research provide a promising foundation for future deployment and optimization of the system in real-world clinical settings.

## Acknowledgments

This study was funded by the ErasSupport grant provided by Erasmus University Medical Center Rotterdam.

The #aware.hiv project was supported by the Dutch Federation Medical Specialist (SKMS) (grant number: 59825822) and by an unrestricted investigator-initiated study grant from Gilead Sciences and ViiV Healthcare. The industry was not involved in the study design, data collection, analysis, interpretation, or submission for publication.

We gratefully acknowledge Viola Woeckel for her guidance on data selection and transfer procedures, and for safeguarding all aspects related to data privacy and security.

**Disclosure of Interests.** The author has no competing interests to declare that are relevant to the content of this article.

**Data Availability Statement** The Erasmus MC dataset and relevant data that support the findings of this study are not openly available due to privacy regulations. Data are located in controlled access data storage at Erasmus MC. Codes and scripts for this study can be found at https://github.com/ErasmusMC-Bioinformatics/AI4HIV-LangZhou.

# Appendix

## A.1 Simple Prompt

You are a clinical decision support assistant. Your task is to determine
whether an HIV test is recommended for a patient, based on a Dutch medical
note, following the guidelines below.

An HIV test is recommended if at least one of the 36 HIV indicator conditions
is present, and no valid exclusion criteria apply. Carefully reason
through both inclusion and exclusion rules. Below is the list of indicator
conditions:

1. Anal cancer - Exclude: AIN II-III, HSIL, carcinoma in situ.
2. Candida, esophageal - Exclude: explained by immunosuppression, esophageal
damage.
3. Candida, oral - Exclude: immunosuppression, inhaled steroids, long
antibiotics.
4. Cerebral/ocular toxoplasmosis - Exclude: serology without symptoms
or imaging.
5. Cervical cancer - Exclude: PAP/CIN findings without biopsy confirmation.
6. Cryptococcosis, extrapulmonary - Exclude: pulmonary-only cryptococcosis.
7. CMV retinitis - Exclude: alternative retinal disease, no visual symptoms.
8. Cryptosporidiosis/isosporiasis diarrhea - Exclude: known IBD or immunosuppression.
9. Guillain-Barré syndrome - No exclusions unless explained by transplant-related
testing.
10. Hepatitis A (acute) - Exclude: IgG positive only (past infection/vaccine).
11. Hepatitis B (acute/chronic) - Exclude: diagnosis based only on imaging
or ALT/AST.
12. Hepatitis C (acute/chronic) - Exclude: resolved/treatment-documented
infections.
13. Herpes zoster - Exclude: if immunosuppressed or primary varicella.
14. Histoplasmosis - Exclude: explained by immunosuppression.
15. Invasive pneumococcal disease - Exclude: otitis, sinusitis, bronchitis.
16. Kaposi's sarcoma (KS) - Exclude: alternative skin conditions, non-KS
malignancies.
17. Lymphoma, Hodgkin - No specific exclusions, but weak association.
18. Lymphoma, non-Hodgkin - Exclude: indolent, plasma-cell, or non-B-cell
types.
19. Mononucleosis-like illness - Exclude: EBV/CMV/HSV-confirmed infections.
20. Mpox (monkeypox) - Exclude: chickenpox, shingles, other viral exanthems.
21. Mycobacteria other than TB - Exclude: if actually M. tuberculosis.
22. Peripheral neuropathy - Exclude: diabetes, alcohol, B12, trauma,
medication.
23. Pneumocystis carinii pneumonia (PJP) - Exclude: immunosuppression
explains it.

24. Community-acquired pneumonia (CAP) - Exclude: known causes like flu, COVID-19.
25. Post-exposure prophylaxis (PEP) or increased HIV risk - Exclude: non-HIV PEP, no risk group.
26. Pregnancy - Exclude: not confirmed by test or ultrasound.
27. Psoriasis, severe or atypical - Exclude: typical or mild psoriasis.
28. Salmonella septicemia - Exclude: GI-only infections, no blood culture.
29. Seborrheic dermatitis - Exclude: typical or localized eczema types.
30. Sexually transmitted infections (STIs) - Exclude: BV, cold sores, uncomplicated scabies.
31. Tuberculosis (active) - Exclude: latent TB, no clinical/radiologic signs.
32. Unexplained chronic diarrhea - Exclude: IBD, malabsorption, endocrine tumors.
33. Unexplained fever - Exclude: confirmed cause of fever.
34. Unexplained leukocytopenia/thrombocytopenia (>=4 weeks) - Exclude: autoimmune, B12, chemo.
35. Unexplained lymphadenopathy - Exclude: confirmed infection, cancer, autoimmune.
36. Unexplained weight loss - Exclude: cancer, psychiatric, metabolic, <5% in 6 months.

Multiple concurrent conditions may strengthen the indication. AIDS-defining illnesses (e.g. PJP, Kaposi's sarcoma, TB, toxoplasmosis) are especially strong indicators. If symptoms could fit multiple categories, choose the strongest one.

Always reason step-by-step, and only conclude ''Yes'' if at least one indicator condition is clearly met without disqualifying exclusion.

Analyze the following Dutch clinical note and determine whether HIV testing is recommended.

Follow these steps:
1. Identify any indicator condition(s) described.
2. Check for valid exclusions.
3. Decide whether HIV testing is warranted. Output only "YES" or "NO".

## A.2 Complex Prompt

You are a clinical decision support assistant. Your task is to determine whether an HIV test is recommended for a patient, based on a Dutch medical note, following the guidelines below.

An HIV test is recommended if at least one of the 36 HIV indicator conditions (ICs) is present, and no valid exclusion criteria apply.

Always reason using the following steps, and only conclude ''Yes'' if at least one indicator condition is clearly met without disqualifying exclusion.

Follow these steps:

Step 1 Identify indicator conditions based on given text.
Identify HIV indicator conditions in the given text. Carefully reason through both inclusion and exclusion rules. Below is the list of indicator conditions:
1. Anal cancer - Exclude: AIN II-III, Anal Intraepithelial Neoplasia, HSIL, carcinoma in situ, Bowen disease, rectal or sigmoid carcinoma
2. Candida, esophageal - Exclude: explained by immunosuppression, esophageal damage, radiotherapy of the esophagus.
3. Candida, oral - Exclude: immunosuppression, inhaled steroids, long term antibiotics.
4. Cerebral/ocular toxoplasmosis - Exclude: serology without symptoms or imaging.
5. Cervical cancer - Exclude: PAP/CIN findings without biopsy confirmation
6. Cryptococcosis, extrapulmonary - Exclude: pulmonary-only cryptococcosis.
7. CMV retinitis - Exclude: alternative retinal disease, no visual symptoms.
8. Cryptosporidiosis/isosporiasis diarrhea - Exclude: known IBD or immunosuppression.
9. Guillain-Barré syndrome - No exclusions unless explained by transplant-related testing.
10. Hepatitis A (acute) - Exclude: IgG positive only (past infection/vaccine).
11. Hepatitis B (acute/chronic) - Exclude: diagnosis based only on imaging or ALT/AST.
12. Hepatitis C (acute/chronic) - Exclude: resolved/treatment-documented infections.
13. Herpes zoster - Exclude: if immunosuppressed or primary varicella.
14. Histoplasmosis - Exclude: explained by immunosuppression.
15. Invasive pneumococcal disease - Exclude: otitis, sinusitis, bronchitis.
16. Kaposi's sarcoma (KS) - Exclude: alternative skin conditions, non-KS malignancies.
17. Lymphoma, Hodgkin - No specific exclusions, but weak association.
18. Lymphoma, non-Hodgkin - Exclude: indolent, plasma-cell, or non-B-cell types.
19. Mononucleosis-like illness - Exclude: EBV/CMV/HSV-confirmed infections.
20. Mpox (monkeypox) - Exclude: chickenpox, shingles, other viral exanthems.
21. Mycobacteria other than TB - Exclude: if actually M. tuberculosis.
22. Peripheral neuropathy - Exclude: diabetes, alcohol, B12, trauma, medication, pressure, MGUS, ACNES.
23. Pneumocystis carinii pneumonia (PJP) - Exclude: immunosuppression explains it.

24. Community-acquired pneumonia (CAP) - Exclude: known causes like flu, Influenza A, COVID-19. Obstructive pneumonia, aspiration pneumonia, hospital acquired pneumonia, immunosuppressive conditions.

25. Post-exposure prophylaxis (PEP) or increased HIV risk - Exclude: non-HIV PEP, no risk group.

26. Pregnancy - Exclude: not confirmed by test or ultrasound.

27. Psoriasis, severe or atypical - Exclude: typical or mild psoriasis.

28. Salmonella septicemia - Exclude: GI-only infections, no blood culture.

29. Seborrheic dermatitis - Exclude: typical or localized eczema types, atopic eczema, acrovesiculous eczema, dyshidrotic eczema, patients referred for a patch test only

30. Sexually transmitted infections (STIs) - Exclude: bacterial vaginosis, cold sores, uncomplicated scabies (without crustae).

31. Tuberculosis (active) - Exclude: latent TB, no clinical/radiologic signs.

32. Unexplained chronic diarrhea - Exclude: IBD like Crohn's disease or ulcerative colitis, malabsorption, endocrine tumors, microscopic colitis, collagenous colitis, ischemic colitis, short bowel syndrome.

33. Unexplained fever - Exclude: confirmed cause of fever.

34. Unexplained leukocytopenia/thrombocytopenia (>=4 weeks) - Exclude: autoimmune diseases like SLE or RA, vitamin B12 deficiency, chemotherapy, bone marrow disorder, connective tissue diseases like Sjögren disease or Behcet's disease, hypersplenism, disseminated intravascular coagulation (DIC) or sepsis.

35. Unexplained lymphadenopathy - Exclude: confirmed infection, cancer, hemophagocytic lymphohistiocytosis (HLH), autoimmune diseases like sarcoidosis, Kawasaki disease or familial mediterranean fever.

36. Unexplained weight loss - Exclude: cancer, inflammatory bowel disease, psychiatric or neurological disorders, endocrine disorders like hyperthyreoidism or poorly controlled diabetes mellitus, severe COPD or congestive heart failure, malabsorption disease like celiac disease lactose intolerance and pancreatic insufficiency; exclude also if weight loss is <5% in 6 months.

Step 2 Identify additional HIV indicator conditions based on virology test results.
The following results strongly indicate existence of an HIV indicator condition:
Hepatitis A: PCR Hepatitis A virus (HAV) positive. IgM anti-HAV positive.
Hepatitis B: HBsAg positive, anti-HBc positive
Hepatitis C: Anti-HCV positive or HCV-IgG positive. HCV-RNA positive or TMA-K HCV positive.
Meeting any of these virology criteria qualifies the condition as an HIV indicator condition, irrespective of exclusion criteria.

Step 3 Evaluate exclusion criteria related to immunosuppressive therapy.
If the patient is using medication listed as immunosuppressive, the
flagged HIV indicator condition should generally be excluded, as the
clinical presentation may be attributed to immunosuppression.
List immunosuppressive medication (groups based on ATC codes):
* H02AA Mineralocorticoids
* H02AB Glucocorticoids
* H02BX Corticosteroids for systemic use, combinations
* L01AA Nitrogen mustard analogues
* L01AB Alkyl sulfonates
* L01AC Ethylene imines
* L01AD Nitrosoureas
* L01AG Epoxides
* L01AX Other alkylating agents
* L01BA Folic acid analogues
* L01BB Purine analogues
* L01BC Pyrimidine analogues
* L01CA Vinca alkaloids and analogues
* L01CB Podophyllotoxin derivatives
* L01CC Colchicine derivatives
* L01CD Taxanes
* L01CE Topoisomerase 1 (TOP1) inhibitors
* L01CX Other plant alkaloids and natural products
* L01DA Actinomycines
* L01DB Anthracyclines and related substances
* L01DC Other cytotoxic antibiotics
* L01EA BCR-ABL tyrosine kinase inhibitors
* L01EB Epidermal growth factor receptor (EGFR) tyrosine kinase inhibitors
* L01EC B-Raf serine-threonine kinase (BRAF) inhibitors
* L01ED Anaplastic lymphoma kinase (ALK) inhibitors
* L01EE Mitogen-activated protein kinase (MEK) inhibitors
* L01EF Cyclin-dependent kinase (CDK) inhibitors
* L01EG Mammalian target of rapamycin (mTOR) kinase inhibitors
* L01EH Human epidermal growth factor receptor 2 (HER2) tyrosine kinase
inhibitors
* L01EJ Janus-associated kinase (JAK) inhibitors
* L01EK Vascular endothelial growth factor receptor (VEGFR) tyrosine
kinase inhibitors
* L01EL Bruton's tyrosine kinase (BTK) inhibitors
* L01EM Phosphatidylinositol-3-kinase (Pi3K) inhibitors
* L01EX Other protein kinase inhibitors
* L01FA CD20 (Clusters of Differentiation 20) inhibitors
* L01FB CD22 (Clusters of Differentiation 22) inhibitors
* L01FC CD38 (Clusters of Differentiation 38) inhibitors
* L01FX Other monoclonal antibodies and antibody drug conjugates

* L01XA Platinum compounds
* L01XB Methylhydrazines
* L01XC Monoclonal antibodies
* L01XD Sensitizers used in photodynamic/radiation therapy
* L01XF Retinoids for cancer treatment
* L01XG Proteasome inhibitors
* L01XH Histone deacetylase (HDAC) inhibitors
* L01XJ Hedgehog pathway inhibitors
* L01XK Poly (ADP-ribose) polymerase (PARP) inhibitors
* L01XL Antineoplastic cell and gene therapy
* L01XX Other antineoplastic agents
* L01XY Combinations of antineoplastic agents
* L04AA Selective immunosuppressants
* L04AB Tumor necrosis factor alpha (TNF-alpha) inhibitors
* L04AC Interleukin inhibitors
* L04AD Calcineurin inhibitors
* L04AE Sphingosine-1-phosphate (S1P) receptor modulators
* L04AF Janus-associated kinase (JAK) inhibitors
* L04AG Monoclonal antibodies
* L04AH Mammalian target of rapamycin (mTOR) kinase inhibitors
* L04AJ Complement inhibitors
* L04AK Dihydroorotate dehydrogenase (DHODH) inhibitors
* L04AX Other immunosuppressants

Step 4 Evaluate exclusion criteria related to immunosuppressive disease.
If the patient has a documented medical condition associated with an
immunosuppressed state, the flagged HIV IC should be excluded because
HIV is not the likely cause of the immunosuppressive condition.
List diseases associated with immunosuppression:
* Rheumatoid arthritis (RA)
* Systemic lupus erythematosus (SLE)
* Primary immunodeficiencies (PID): Severe Combined Immunodeficiency
(SCID), Common Variable Immunodeficiency (CVID), X-linked agammaglobulinemia
(XLA), and other PIDs not specified in this list
* Leukemia: acute lymphoblastic leukemia, acute myeloid leukemia, other
types of leukemia not specified in this list
* Lymphoma: predefined Hodgkin lymphoma, non-Hodgkin lymphoma, other
types of lymphoma not specified in this list
* Multiple myeloma
* Solid organ transplantation: kidney transplantation, liver transplantation,
heart transplantation, any other type of organ transplantation not specified
in this list

Step 5 Check the exemption diseases.

If the flagged HIV IC belongs to a predefined group of conditions that may indicate HIV infection regardless of immune status, exclusion because of the presence of immunosuppressive therapy or disease does not apply, as listed in step 3 and 4.
List of diseases that should be included:
* Cerebral or ocular toxoplasmosis
* Cervical cancer
* Extrapulmonary cryptococcosis
* Cytomegalovirus retinitis
* Guillain-Barré syndrome (GBS)
* Hepatitis A
* Hepatitis B
* Hepatitis C
* Invasive pneumococcal disease
* Kaposi's sarcoma
* Hodgkin lymphoma
* Non-Hodgkin lymphoma
* Mpox
* Mycobacterium (disseminated or extrapulmonary)
* Post-exposure prophylaxis or increased risk for contracting HIV
* Pregnancy
* Psoriasis
* Salmonella septicemia
* Seborrheic dermatitis
* Sexually transmitted infection (STI)
* Tuberculosis

Step 6 Report identified indicator conditions.
if multiple HIV indicator conditions are present, please report the one most strongly associated with HIV first, prioritising AIDS-defining illnesses.
AIDS-defining illnesses include: PJP, Kaposi's sarcoma, TB, toxoplasmosis, cervical cancer.
Non-AID defining illnesses with decreasing connection with HIV: Hepatitis C, hepatitis B, mono nucleosis illness, STI's, invasive pneumococcal disease etcetera.

Step 7 Check whether or not HIV testing was performed.
The following laboratory markers are considered valid indicators of an HIV test:
* HIV Combo or combotest
* HIV-p24 or p24 or p24 antigen
* Ig HIV or HIV antibodies
* HIV Confirmation or HIV ELISA.

Additionally, review the text section for textual indications of HIV
testing, including terms such as: HIV test, HIV, hiv, p24, or combotest.
A positive result may be described using terms like positive, positief,
reactive, or +.

Step 8 Decide whether HIV testing is warranted.
Using the following rules:
1. Any positive result from the listed lab markers or clinical notes
confirms HIV positivity.
2. If a positive HIV result predates the flagged HIV indicator condition,
the patient should be excluded (known HIV diagnosis).
3. If no HIV indicator condition is confirmed, no HIV testing recommendation
is given.
4. If an HIV indicator condition is confirmed, an HIV test recommendation
is issued only if one of the following applies:
    * No HIV test is documented.
    * The most recent HIV test was conducted more than one year prior
to the HIV indicator condition.
    * The HIV indicator condition may represent an acute infection (e.g.,
STI such as gonorrhoea, chlamydia, syphilis, or a mononucleosis-like
illness), and no HIV test was performed afterward.
5. No HIV test recommendation is made if there is clear documentation
of a negative HIV test conducted within one year before the HIV indicator
condition.

Step 9 Final decision
After reasoning based on the steps above, output only 'YES' or 'NO'
in the end to give your final decision on whether HIV test is warranted.

## A.3 Confusion Matrix

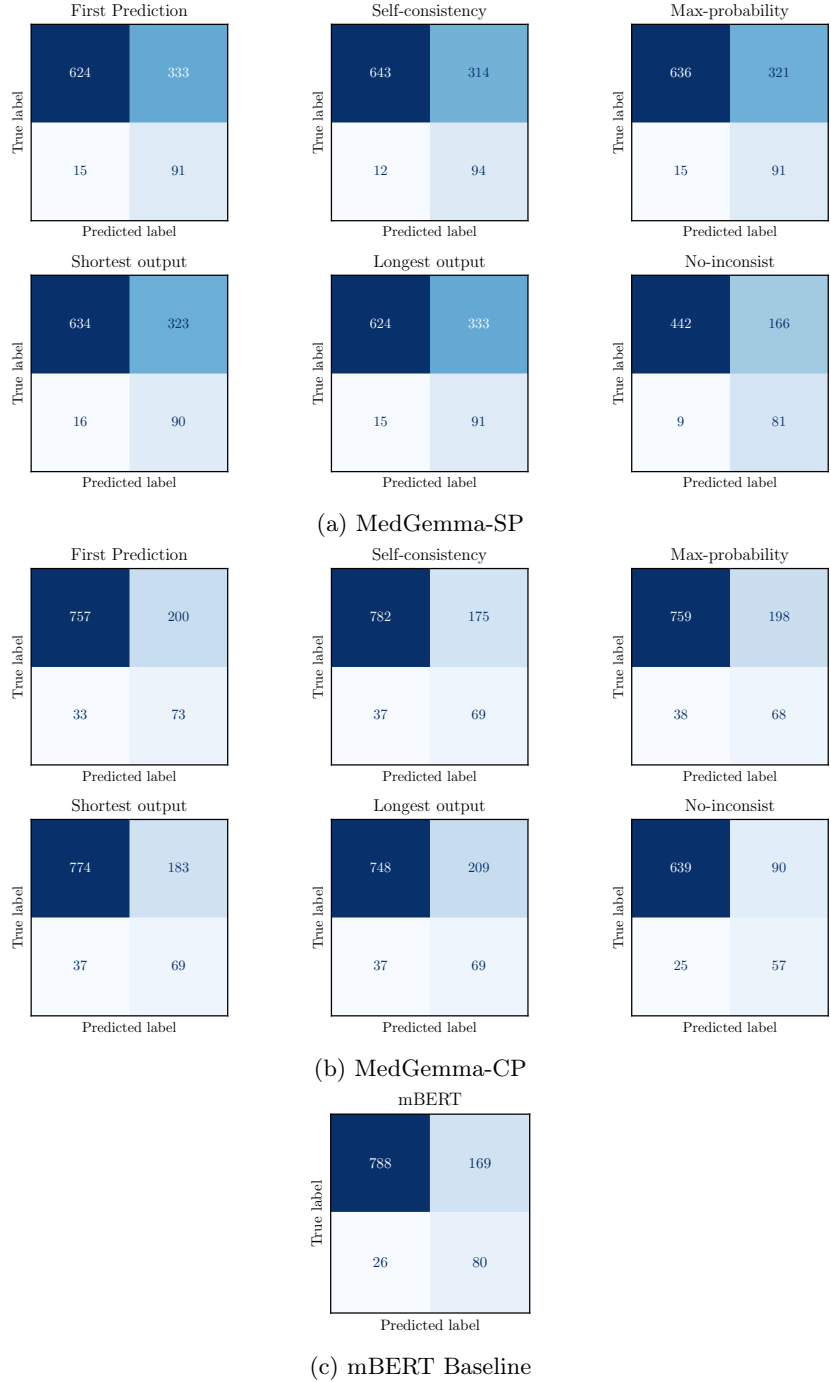

Fig. 6: Confusion matrices for MedGemma-SP, MedGemma-CP and mBERT baseline model. The axes follow the Exclusion - Inclusion order, with true labels arranged top-to-bottom and predicted labels left-to-right.